\documentclass[final]{cvpr}

\usepackage{times}
\usepackage{epsfig}
\usepackage{graphicx}
\usepackage{amsmath}
\usepackage{amssymb}
\usepackage[utf8]{inputenc} %
\usepackage[T1]{fontenc}    %

\usepackage{amsmath,amsfonts,bm}

\def\eqref#1{equation~\ref{#1}}

\def\1{\bm{1}}

\def\vs{{\bm{s}}}

\DeclareMathAlphabet{\mathsfit}{\encodingdefault}{\sfdefault}{m}{sl}
\SetMathAlphabet{\mathsfit}{bold}{\encodingdefault}{\sfdefault}{bx}{n}

\usepackage{color,xcolor}
\usepackage{epsfig}
\usepackage{adjustbox}
\usepackage{array}
\usepackage{booktabs}
\usepackage{colortbl}
\usepackage{wrapfig}
\usepackage{hhline}
\usepackage{multirow}

\usepackage{paralist}
\usepackage{tabularx}

\usepackage{amsmath,amsfonts,amssymb}
\usepackage{bm}
\usepackage{nicefrac}
\usepackage{microtype}

\usepackage{changepage}
\usepackage{extramarks}
\usepackage{fancyhdr}
\usepackage{lastpage}
\usepackage{setspace}
\usepackage{soul}
\usepackage{xspace}

\usepackage{tabularx}

\usepackage[pagebackref=true,breaklinks=true,colorlinks,bookmarks=false]{hyperref}

\usepackage{url}

\usepackage{enumerate}
\usepackage{enumitem}  %

\usepackage{makecell}

\usepackage{pifont} %
\usepackage[ruled,vlined]{algorithm2e}
\usepackage{tabularx} %
\newcolumntype{L}[1]{>{\raggedright\let\newline\\\arraybackslash\hspace{0pt}}m{#1}}
\newcolumntype{C}[1]{>{\centering\let\newline\\\arraybackslash\hspace{0pt}}m{#1}}
\newcolumntype{R}[1]{>{\raggedleft\let\newline\\\arraybackslash\hspace{0pt}}m{#1}}

\newcommand{\fig}[1]{Figure~\ref{#1}}
\newcommand{\tbl}[1]{Table~\ref{#1}}

\newcommand{\ignore}[1]{}

\makeatletter
\DeclareRobustCommand\onedot{\futurelet\@let@token\@onedot}
\def\@onedot{\ifx\@let@token.\else.\null\fi\xspace}

\def\eg{e.g\onedot} 
\def\ie{i.e\onedot}

\def\vs{\emph{vs}\onedot}

\makeatother

\definecolor{MyDarkBlue}{rgb}{0,0.08,1}
\definecolor{MyDarkGreen}{rgb}{0.02,0.6,0.02}
\definecolor{MyDarkRed}{rgb}{0.8,0.02,0.02}
\definecolor{MyDarkOrange}{rgb}{0.40,0.2,0.02}
\definecolor{MyPurple}{RGB}{111,0,255}
\definecolor{MyRed}{rgb}{1.0,0.0,0.0}
\definecolor{MyGold}{rgb}{0.75,0.6,0.12}
\definecolor{MyDarkgray}{rgb}{0.66, 0.66, 0.66}

\newcommand{\myparagraph}[1]{\vspace{-12pt}\paragraph{#1}}
\newcommand{\myitem}{\vspace{-0pt}\item}

\newcommand{\Model}{Motion Programs\xspace}
\newcommand{\model}{motion programs\xspace}

\def\cvprPaperID{5568} %
\def\confYear{CVPR 2021}

\begin{document}

\title{Hierarchical Motion Understanding via Motion Programs}

\author{Sumith Kulal\thanks{Equal contribution. Email: {\tt sumith@cs.stanford.edu}. \newline Webpage: \url{https://sumith1896.github.io/motion2prog}.}\\
Stanford University\\
\and
Jiayuan Mao\footnotemark[1]\\
MIT \\
\and
Alex Aiken\\
Stanford University\\
\and
Jiajun Wu \\
Stanford University\\
}

\maketitle
\pagestyle{empty}
\thispagestyle{empty}

\begin{abstract}
\vspace{-1em}
Current approaches to video analysis of human motion focus on raw pixels or keypoints as the basic units of reasoning.  We posit that adding higher-level motion primitives, which can capture natural coarser units of motion such as ``backswing'' or ``follow-through'', can be used to improve downstream analysis tasks. This higher level of abstraction can also capture key features, such as loops of repeated primitives, that are currently inaccessible at lower levels of representation.  We therefore introduce \emph{\Model}, a neuro-symbolic, program-like representation that expresses motions as a composition of high-level primitives. We also present a system for automatically inducing motion programs from videos of human motion and for leveraging motion programs in video synthesis. 
Experiments show that motion programs can accurately describe a diverse set of human motions and the inferred programs contain semantically meaningful motion primitives, such as arm swings and jumping jacks. Our representation also benefits downstream tasks such as video interpolation and video prediction and outperforms off-the-shelf models. We further demonstrate how these programs can detect diverse kinds of repetitive motion and facilitate interactive video editing. 
\end{abstract}

\section{Introduction}

Most current video analysis architectures operate on either pixel-level or keypoint-level transitions that are good at local and short-term motions of human bodies but do not handle higher-level spatial and longer-lasting temporal structures well. 
We posit that incorporating higher-level reasoning of motion primitives and their relationships enables better video understanding and improved downstream task performance: after seeing a person perform jumping jacks a few times, it's reasonable to predict that they may repeat the same action for a few more iterations. %

\begin{figure*}[t]
    \centering
    \includegraphics[width=\textwidth]{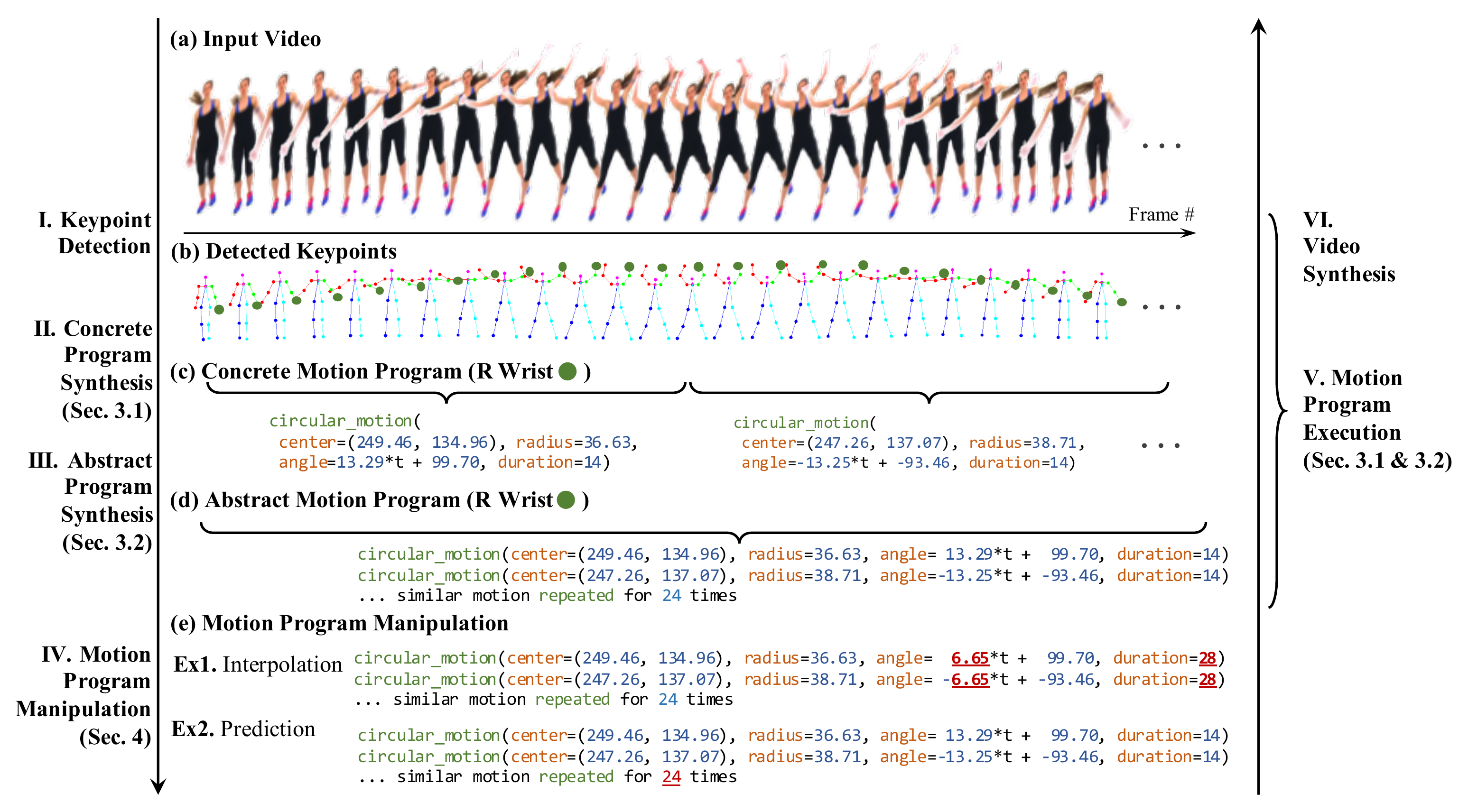}
    \caption{Overview of our hierarchical motion understanding framework. We start with low-level keypoints extracted from raw videos and build higher level motion understanding. Higher levels of abstraction consist of a sequence of motion primitives and their repetitive patterns. Similar motion in repetitive patterns are captured using a probabilistic for-loop body (refer Figure~\ref{fig:loop-roll}).} 
    \label{fig:framework}
\end{figure*}

We thus introduce a hierarchical motion understanding framework with three levels of abstraction, as shown in \fig{fig:framework}:
\begin{compactitem}
    \myitem Level 1 (\fig{fig:framework}b) is \textbf{Keypoints}, which are the basis of most current computer vision systems. This level provides low-level priors such as temporal smoothing.
    \myitem Level 2 (\fig{fig:framework}c) is \textbf{Concrete Motion Programs}, a programmatic representation of motion consisting of a sequence of atomic actions or \emph{motion primitives}, each describing a set of frames for a single action. This level provides priors on primitive level motion.
    \myitem Level 3 (\fig{fig:framework}d) is \textbf{Abstract Motion Programs}, which describe higher-level groupings of the concrete primitives, such as patterns of repetitive motion. This level provides priors on sequences of primitives and their patterns.
\end{compactitem}

We describe the concrete and abstract levels with simple domain specific languages (DSLs). The Concrete DSL consists of motion primitives such as stationary, linear, and circular movement with necessary spatial and temporal motion parameters. The Abstract DSL includes for-loops for capturing higher-level groupings of primitives. Along with the DSLs, we propose an algorithm to automatically synthesize \model from videos. For the sake of brevity, we often refer to \model as simply programs.

To demonstrate the advantages of motion programs, we apply our methods for synthesizing motion programs from videos to three applications: video interpolation (Section~\ref{sec:exp-interpolate}), repetitive segment detection (Section~\ref{sec:exp-loop}) and video prediction (Section~\ref{sec:exp-manip}).
For video interpolation, we demonstrate that when combined with state-of-the-art image synthesis models, such as SepConv~\cite{sepconv}, \model can achieve higher quality interpolations with bigger gaps between frames than previous approaches.  For repetitive segment detection, we demonstrate that  \model can detect and extract repetitive segments of human motion. For video prediction, we show that \model can perform longer-range video prediction than HierchVid~\cite{villegas17hierchvid} and HistryRep~\cite{wei2020his} by identifying repetitive motions.

To summarize, our contributions are:
\begin{compactitem}
    \myitem We introduce a hierarchical motion understanding framework using \model.
    \myitem  We present an algorithm for automatically synthesizing motion programs from raw videos.
    \myitem By incorporating motion programs into video synthesis pipelines, we demonstrate higher performance in video interpolation, video prediction and also show how to detect and extract repetitive motion segments. 
\end{compactitem}

\section{Related Work}
\label{sec:related}

We briefly summarize the most relevant related work in capturing and understanding images and video.

\myparagraph{Program-like representations for visual data.} 
The idea of using programs to represent visual data dates back to the early computer graphics literature on using procedural models to model biological structures~\cite{lindenmayer1968mathematical}, 3D shapes~\cite{li2017grass,sharma2018csgnet,tian2019learning}, and indoor scenes~\cite{wang2011symmetry,li2019grains,niu2018im2struct}. Recent works integrate the representational power of deep networks with program-like structures in visual data modeling, which enables the application of high-level program-like structures in more complex and noisy environments such as hand-drawn sketches~\cite{ellis2018learning} and even natural images~\cite{young2019learning,mao2019program}. These works have all focused on extracting programmatic descriptions of images, whereas our main contribution is in developing and applying programmatic descriptions of motion to videos.

Our model also relates to works on using parameterized curves~\cite{rose1999verbs,gleicher1997motion,gleicher1998constraint} or motion segments that can be recombined~\cite{kovar2008motion,aristidou2018deep} for modeling and editing motions. Compared with these works,  we focus on inferring program-like representation of human motion from raw videos, and introduce abstract motion programs to describe higher-level groupings of low-level motions, which enables us to capture patterns of repetitive motion. As a result, the applications of such representations are different, with our programmatic representation being useful for multiple video manipulation tasks.

\myparagraph{Video synthesis.} Substantial progress has recently been made on video processing tasks including video interpolation~\cite{liu2017video,jiang2018super,liu2019deep,sepconv}, video extrapolation~\cite{villegas17hierchvid,denton2018stochastic}, and video retiming~\cite{lu2020layered,davis2018visual}.
These methods use deep convolutional neural networks (CNNs) to extract spatial-temporal features of videos. These features are sent to task-specific networks to make predictions of intermediate frames (in video interpolation) or future frames (in video extrapolation). Various methods have been proposed to improve the results by predicting flows~\cite{liu2019deep}, predicting blending kernels~\cite{sepconv}, making future predictions in keypoint spaces~\cite{villegas17hierchvid}, and using variational inference to model video stochasticity~\cite{denton2018stochastic}. Our approach
works at a higher level of abstraction and captures structure on longer time scales.  We show that the two classes of work can be combined to yield improved results for video interpolation.

\section{Hierarchical Motion Understanding}

Our hierarchical motion understanding framework has three levels of increasing abstraction. The first level is representing motion as a sequence of keypoints. For the next two levels, we use programmatic representations of motion via two simple domain-specific languages (DSLs). 

In Section \ref{sec:concrete-synth}, we describe the synthesis and execution of {\em concrete motion programs}  (Level 2). In Section \ref{sec:abstract-synth}, we describe synthesis and execution of {\em abstract motion programs}  (Level 3). 

\myparagraph{Level 1---Keypoints.} Representing human motion as a sequence of keypoints is a well-studied problem. Several high-quality, off-the-shelf tools such as AlphaPose \cite{alphapose} and OpenPose \cite{openpose} take raw video as input and output a sequence of 2D keypoints for each body joint.  Sequences
of keypoints for a single joint are our first level of abstraction.

\myparagraph{Level 2---Concrete Motion Programs.} A concrete motion program is a sequence of motion primitives, where each primitive is either circular, linear, or stationary (not moving) motion of a specific keypoint for a fixed time duration. \tbl{tab:concrete-dsl} gives a grammar for the DSL. These programs are \emph{concrete} because the specific kind of motion (\eg, linear) of each primitive in the sequence is explicit as are all the parameters needed to fully describe the motion.  

\begin{figure*}[t]
    \centering
    \includegraphics[width=\textwidth]{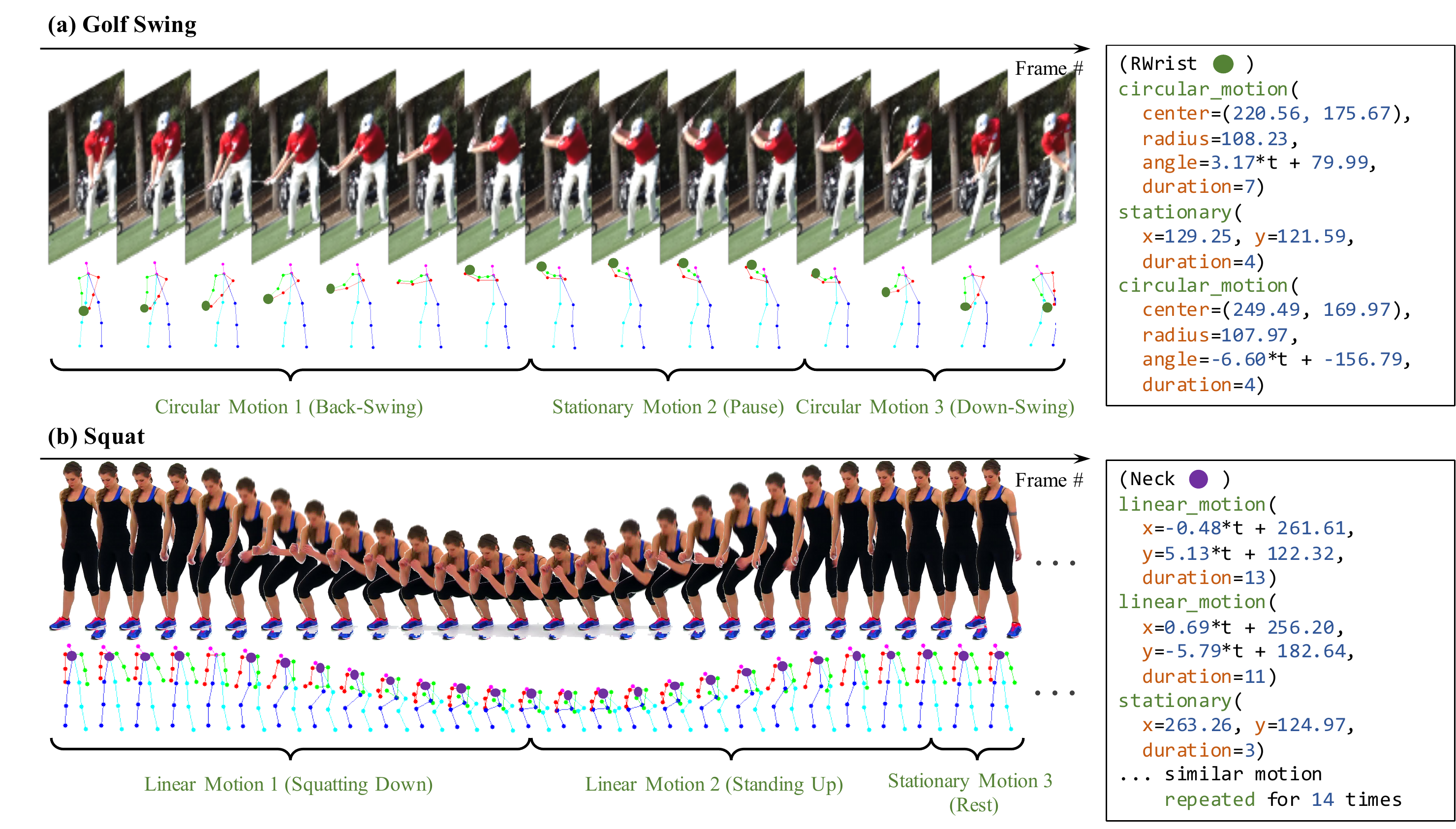}
    \caption{Examples of synthesized motion primitives. a) This golf swing has three primitives: back-swing, pause and down-swing. %
    b) The squats sequence has a repeating subsequence of three similar primitives: squating down, standing up and a brief rest in the standing pose.}
    \label{fig:prim-final}
    \vspace{-10pt}
\end{figure*} 
More specifically, we define three parameterized motion primitives: a circular primitive ({\tt circle}), a linear primitive ({\tt line}) and a stationary primitive ({\tt stationary}). Each primitive has spatial parameters, such as {\tt center} and {\tt radius} for circular motion, temporal motion parameters such as {\tt velocity} and {\tt start\_angle}, and a time duration. In practice,  we have found these three simple motion primitives are sufficient to cover a wide variety of use cases. The framework could easily extended with more primitives or more expressive primitives (such as spline curves).  The execution model of concrete programs is the obvious one: The keypoint traces out each primitive $p$ in order, beginning in the starting configuration for $p$ and for $p$'s specified amount of time.

\myparagraph{Level 3---Abstract Motion Programs.}   Abstract motion programs (see \tbl{tab:abstract-dsl}) consist of sequences and loops of primitive distributions over start, middle, and end points of the motion.  Note that the start-middle-end representation does not commit to the type of motion (\eg, circular \vs linear) and in fact a specific distribution might include instances of multiple kinds of motion.  To execute an abstract primitive $p$, we sample the start, middle and end points (as well as other parameters, such as duration) of $p$'s distribution, pick the concrete primitive $c$ that best fits the sample, and then
execute $c$.  Sequences and loops of abstract primitives are then executed by repeating this process for each abstract primitive in the execution sequence.  The start-middle-end
representation could be extended with more points, but at least three points are needed to distinguish linear from circular motion.

Abstract motion programs concisely capture complex categories of human motion.  For example, ``jumping jacks'' are easily expressed by a loop with a body consisting of a sequence of primitives describing the up and down motion of a single jumping jack.  The fact that the primitives are distributions captures the natural variation that arises in repeated jumping jacks---no two will be exactly the same, but all will be similar within some bounds.

\begin{table}

\setlength{\tabcolsep}{2pt}
    \centering
    \caption{DSL of Concrete Programs (Level 2) for describing motion as a sequence of motion primitives.}
    \label{tab:concrete-dsl}
    \vspace{5pt}
    \begin{tabularx}{\columnwidth}{rcl}
    \toprule
    Program & $\longrightarrow$ & Prim; \dots; Prim \\
    {Prim} & {$\longrightarrow$} & {\tt circle} (center: {\normalsize Point}, radius: {\normalsize R},\\
    & & \quad angle={\normalsize (vel: R, start: R)}, time: {\normalsize Int}) \\
    {Prim} & {$\longrightarrow$} & {\tt line} (x={\normalsize (vel: R, start: R)},\\
    & & \quad y={\normalsize (slope: R, intercept: R)}, time: {\normalsize Int}) \\
    {Prim} & {$\longrightarrow$} & {{\tt stationary}(point: {\normalsize Point}, time: {\normalsize Int})} \\
    {Point} & {$\longrightarrow$} & {(x: {\normalsize R}, y: {\normalsize R})} \\
    {Int} & {$\longrightarrow$} & {a positive integer} \\
    {R} & {$\longrightarrow$} & {a real number}  \\
    \bottomrule
    \end{tabularx}
\end{table} 
\begin{table}[t!]

\setlength{\tabcolsep}{2pt}
    \centering
    \caption{DSL of Abstract Programs  (Level 3) for describing motion as a sequence of probabilistic motion primitives with loops.}
    \vspace{5pt}
    \label{tab:abstract-dsl}
    \begin{tabularx}{\columnwidth}{rcl}
    \toprule
    Program & $\longrightarrow$ & Stmt; \dots; Stmt \\
    Stmt & $\longrightarrow$ & DetPrim $|$ For-Loop (iter: {\normalsize Int}) \{\\
    & & \quad ProbPrim; \dots; ProbPrim \} \\
    {DetPrim} & {$\longrightarrow$} & \{<start, middle, end>: \\
    & &<{\normalsize Point}, {\normalsize Point}, {\normalsize Point}>, time: {\normalsize Int}\} \\
    {ProbPrim} & {$\longrightarrow$} & \{<start, middle, end>: \\
    & & Gaussian(<{\normalsize Point}, {\normalsize Point}, {\normalsize Point}>), \\
    & & ), time: FreqDist({\normalsize Int})\} \\
    {Point} & {$\longrightarrow$} & {(x: {\scriptsize R}, y: {\scriptsize R})} \\
    {Int} & {$\longrightarrow$} & {a positive integer} \\
    {R} & {$\longrightarrow$} & {a real number}  \\
    \bottomrule
    \end{tabularx}
    \vspace{-10pt}
\end{table} 
\subsection{Concrete Motion Programs}
\label{sec:concrete-synth}

We now describe how we connect the different levels of our hierarchy.  Given a sequence of keypoints we synthesize a concrete motion program, and given a concrete motion program we identify repeating patterns in the sequence of concrete primitives to synthesize the distributions and loops of an abstract motion program.  We begin with concrete motion programs.

\myparagraph{Synthesizing single primitives.} To fit a single circular primitive to a sequence of keypoints, we first find the best-fit circle of the sequence \cite{bullock2006least}. We then project the keypoints onto this best-fit curve and find temporal motion parameters ({\tt velocity}, {\tt start\_angle}) that best approximate the motion of the projected points. The problem can also be solved by jointly optimizing for all parameters, but in practice we observed that this two step approach produced equivalent results with significantly faster running times. We run this process for linear and stationary motion and pick the concrete primitive that minimizes the $\mathcal{L}_2$ error.
Some examples of single primitive synthesis are given in Figure \ref{fig:prim-final}.

\myparagraph{Segmentation and synthesis.} A long video is rarely well-described by a single concrete primitive.  To synthesize a sequence of primitives, we must segment the keypoint sequence and synthesize a primitive for each segment.  We use dynamic programming to segment the input sequence at the best possible locations (\ie, the ones that minimize overall error) and fit individual primitives to each segment, as illustrated in Figure \ref{fig:prim-final}. The recurrence relation for the best fit sequence of primitives for the first $n$ keypoints is given as

\begin{equation}
    \mathrm{Error}_{\text{n}} = \min\limits_{k < n} \left[ \mathrm{Error}_{\text{k}} + \mathrm{fit}(\text{keypoints}[k:n]) + \lambda \right].
\end{equation}

\begin{figure*}[t]
    \centering
    \includegraphics[width=\textwidth]{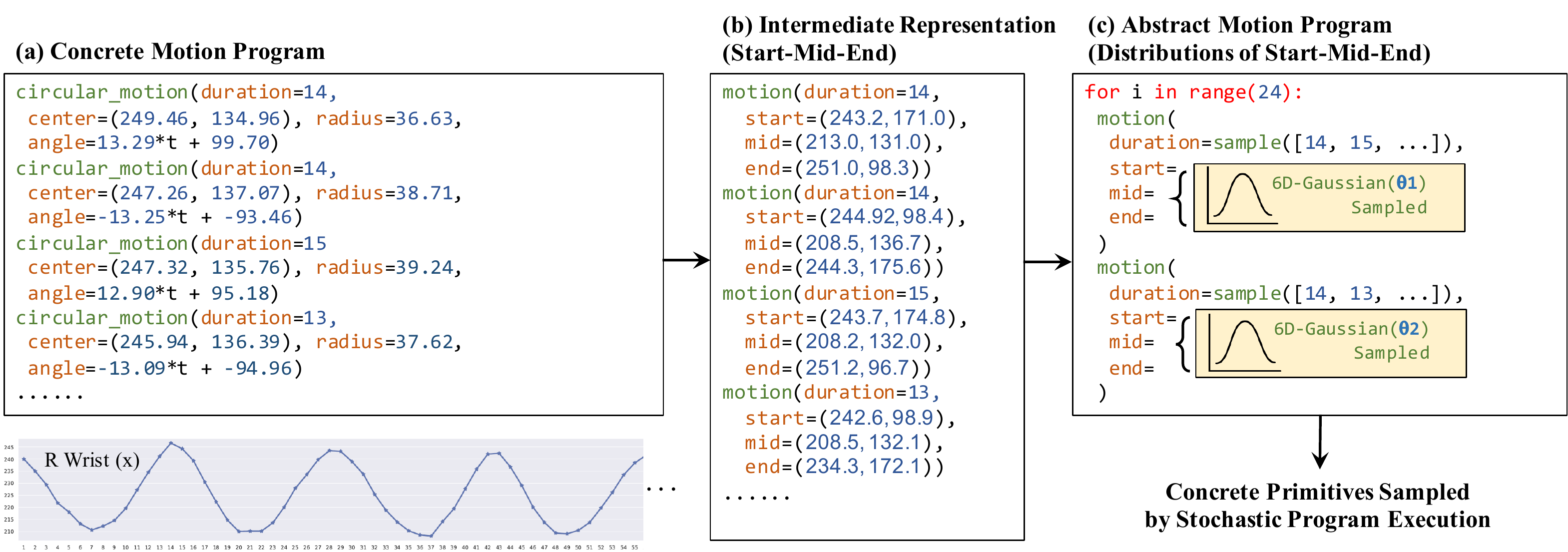}
    \caption{Illustration of rolling up 6 repetitive (alternating) statements into a for-loop of body size 2. We first translate concrete primitives to deterministic abstract primitives and then synthesize for-loops with probabilistic primitives in the body. Concrete primitives are sampled from the probabilistic abstract primitives during execution.}
    \label{fig:loop-roll}
    \vspace{-10pt}
\end{figure*} 
Here, $\mathrm{Error}_{\text{k}}$ is the error incurred by the best fit sequence of primitives for the first $k$ keypoints and $\mathrm{fit}(\text{keypoints}[k:n])$ returns the error incurred by the best-fit single primitive for keypoints from $k$ to $n$. The parameter $\lambda$ is a regularization term that controls the granularity of fitted primitives. Note that $\lambda = 0$ results in a degenerate fit with every point labeled as an independent primitive and $\lambda \to \infty$ results in a single primitive. Hence, in practice, we adaptively choose $\lambda$ at runtime to match the range of the keypoint motion by setting $\lambda$ proportional to the covariance of keypoints in a fixed window centered on the current keypoint.

\myparagraph{Multiple keypoints.} The segmentation and synthesis algorithm can be generalized from one keypoint trajectory to multiple keypoint trajectories by jointly finding the best single segmentation across all the trajectories by minimizing the sum of errors across all keypoints. This approach enforces the useful invariant that programs for all keypoints transition to new primitives simultaneously.  Note that this is the only place where we have found it useful to jointly consider all keypoint trajectories; otherwise motion programs are constructed per keypoint.

Putting it all together, some examples of synthesized primitives after segmentation generated from golf swings and squats are shown in Figure~\ref{fig:prim-final}.

\subsection{Abstract Motion Programs}
\label{sec:abstract-synth} 
The translation from a concrete motion program to an abstract motion program is done in three steps: converting concrete primitives to abstract primitives, loop detection, and loop synthesis.

\myparagraph{Concrete to abstract translation.}   We first replace each concrete primitive with a deterministic abstract primitive. We do this by obtaining the start, middle and end points of the concrete primitive along with the duration of execution.

\myparagraph{Loop detection.} In the second step we detect loops in the sequence of abstract primitives.  We use a sliding window of $W$ primitives and attempt to fit loops with different numbers of abstract primitives $l$ in the loop body for each window. For each candidate loop body size $l$ and $W = \langle w_0,\ldots,w_n\rangle$ we first group together all the deterministic primitives that 
would be in the $i$th abstract primitive of the loop body:
\begin{equation}
S_i = \{ w_j | j\%l = i \} \ \ \mbox{for\ } 0 \leq i < l.     
\end{equation}
Now $S_0,\ldots,S_{l-1}$ is easily converted to a loop by fitting a Gaussian
distribution to each $S_i$ to create an abstract primitive of the loop body.
Note that we could not directly learn such a distribution of parameters for concrete primitives since each set could have more than one primitive type.

We define the quality of fit for a loop as the average of the Gaussian covariance norms across $S_i$. We threshold the quality and below a certain value mark it as a successful loop detection. To find the maximal loop, we dynamically increase the window size to include more primitives till covariance norms are still below threshold. We search over increasing loop body lengths $l$ until either a loop is successfully detected or we exceed a maximum loop body size. 

\myparagraph{Loop synthesis.} As the last step, we replace intervals of abstract primitives with the detected for-loop.
The number of iterations is set according to the number of lines rolled up and the loop body size.
An illustration of translating concrete primitives to abstract primitives and subsequent loop detection is presented in Figure \ref{fig:loop-roll}.

\section{Experiments}

To evaluate the expressive power of both concrete and abstract motion programs, we first evaluate the accuracy of synthesized motion programs (Section~\ref{sec:exp-induction}). Second, we evaluate the motion programs on video interpolation and compare it with state-of-the-art models (Section~\ref{sec:exp-interpolate}). Third, we present results for detecting and extracting repetitive motion from long exercise videos (Section~\ref{sec:exp-loop}). Finally, we demonstrate interactive manipulation of repetitive motion in videos using abstract motion programs and compare it to video prediction baselines (Section~\ref{sec:exp-manip}).

\begin{figure*}[t]
    \centering
    \includegraphics[width=\textwidth]{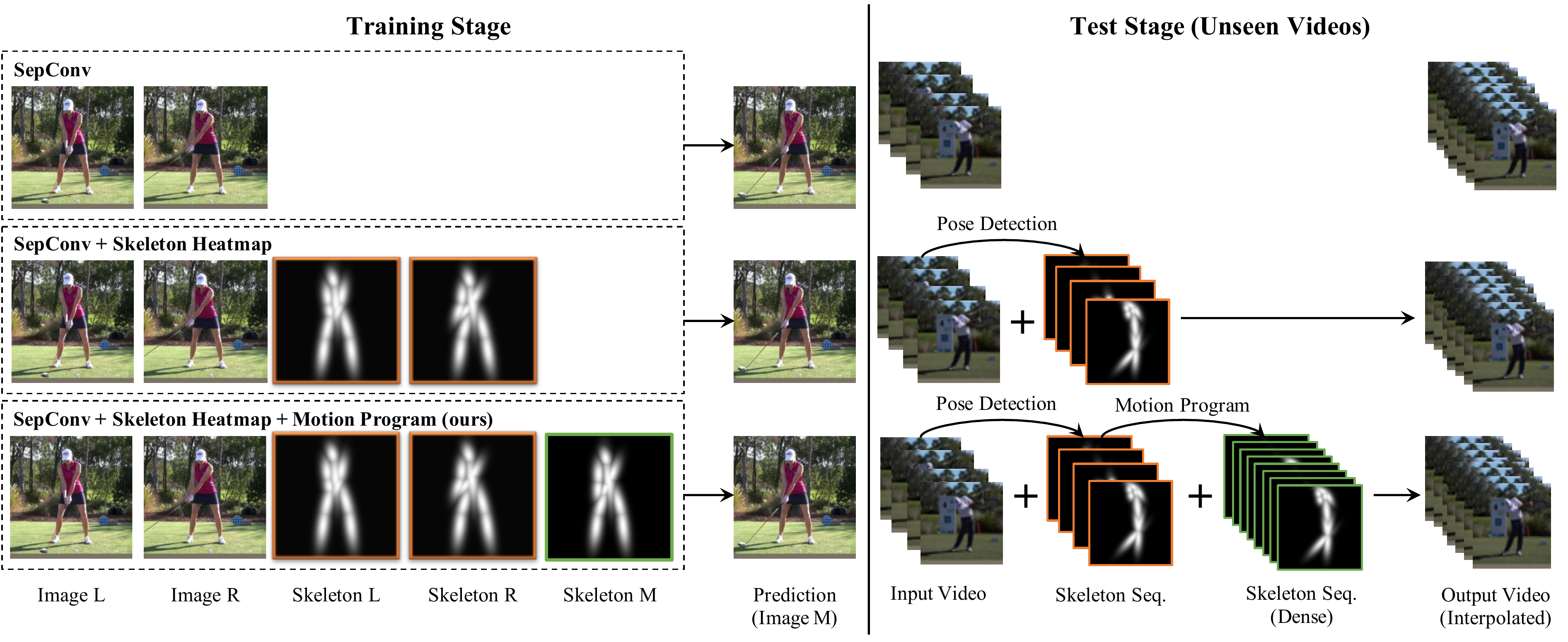}
    \caption{Evaluation of three video interpolation variants: Top row (SepConv \cite{sepconv}), middle row (SepConv w/ heatmaps) and bottom row (SepConv w/ programs) which is our model where poses for intermediate frames are supplied by motion programs. Middle row (SepConv w/ heatmaps) is evaluated as an ablation study.}
    \label{fig:interpolate-train}
    \vspace{-15pt}
\end{figure*}

\subsection{Motion Program Induction}
\label{sec:exp-induction}

\paragraph{Data.} We use the GolfDB dataset~\cite{golfdb} for evaluating the representative power of concrete programs. We extract videos from swing start to swing end from all 242 face-on real-time golf swing videos. We use a 50-50 train-test split and scale each frame to 512$\times$512.

\myparagraph{Analysis.} We synthesize concrete motion programs for the train set of GolfDB and compare the synthesized programs with ground-truth poses. On average, we synthesize 3.33 primitives per video  where the length of videos averages around 58 frames. In the majority of these videos, the primitives semantically correspond to back-swing, down-swing and follow-through with some stationary primitives at the beginning and the end. We analyse the efficiency, accuracy and temporal consistency of program representation. For efficiency, we compare the average number of parameters needed to represent the programs and the input pose sequence. Programs needed 253 parameters per video, which is much lower compared to 1,623 parameters needed by input poses. For accuracy, we compute the keypoint difference metric (KD) of the program-encoded pose sequence with the input pose; KD is the average distance of keypoints with the ground-truth keypoints in pixel space. We observe that the program-encoded poses differ from the ground-truth poses by 6.25 pixels, an error of only 2.44\%. This indicates that motion programs are good approximates of input pose sequence. Due to inherent noise in pose detection, the input poses could at times be noisy. We observe that the motion program representation smoothens out this noise to some degree by. To validate this, we compute the maximum difference in pixels of the  pose of a joint in two adjacent frames. We observe that the input pose sequence has a difference of 100.10 pixels while program-encoded pose sequence are much smoother (76.15 pixels).

\subsection{Video Interpolation}
\label{sec:exp-interpolate}

We leverage motion programs to perform video interpolation and demonstrate the benefit of having primitive-level information in such tasks.

\myparagraph{Data.} We evaluate on two datasets, GolfDB (as described above) and PoseWarp~\cite{posewarp}. For the PoseWarp dataset, we stitch together continuous frames to obtain 105 videos.  We scale each video to 512x512 and use the training splits specified in the original paper.

\myparagraph{Setup.} We take the standard SepConv~\cite{sepconv} architecture and modify it to incorporate the program information as shown in Figure~\ref{fig:interpolate-train}. We train three variants of SepConv: i) SepConv, as defined in the original paper, that predicts the interpolated frame given left and right frames; ii) SepConv with input pose information of both frames (SepConv w/ heatmaps); iii) SepConv with both input and target pose information (SepConv w/ programs). For the last model, at test time, the the target pose information is supplied by a motion program. We synthesize a motion program on input video and generate intermediate poses from executing the program with a finer granularity of time. Note that this information is otherwise not available. We are interested in the first and third model, where the second model is also studied as an ablation. 

For evaluation, we generate programs on the test set of both datasets. For each primitive generated, we sub-sample the frames at different rates (2x/4x/8x) and try to reconstruct the original video by interpolating the remaining frames. Performing 4x interpolation is more challenging than 2x and 8x is even more challenging.

\begin{figure}[t]
    \centering
    \vspace{-10pt}
    \includegraphics[width=\linewidth]{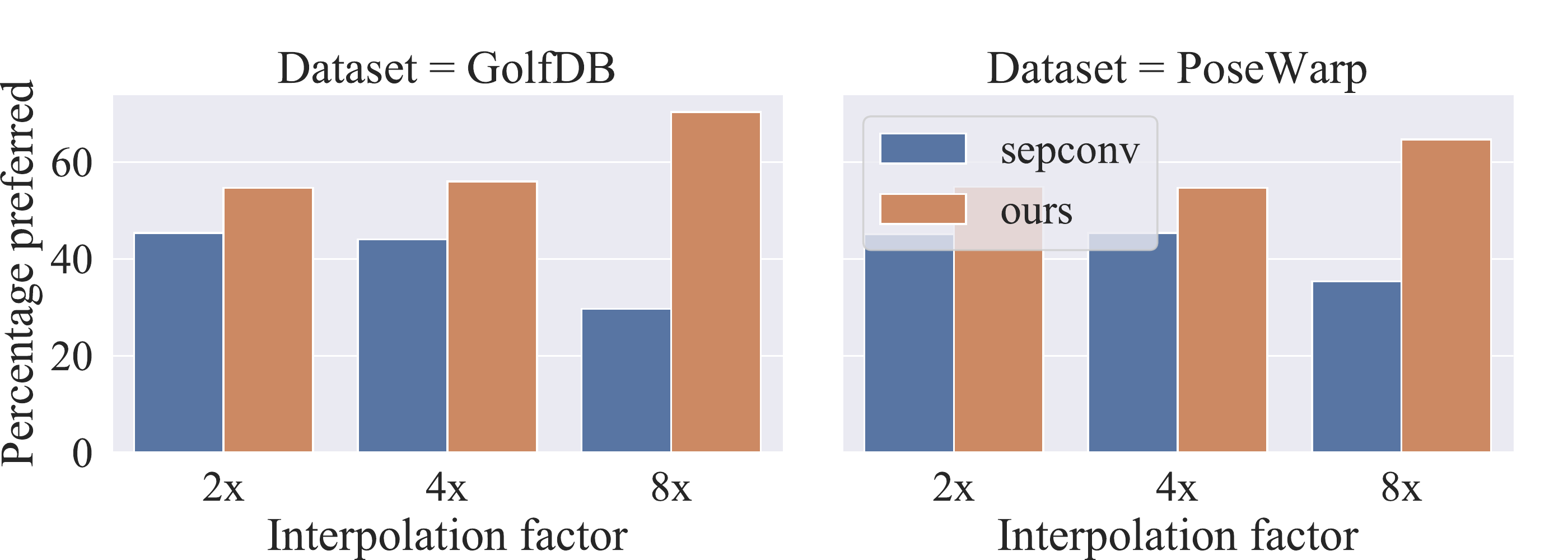}
    \caption{Human evaluation of video interpolation across different rates on both datasets. Incorporating motion programs significantly improves upon the baseline for 8x interpolation.}
    \label{fig:interpolate-human-eval}
    \vspace{-10pt}
\end{figure}
\begin{figure*}[t]
    \centering
    \includegraphics[width=\textwidth]{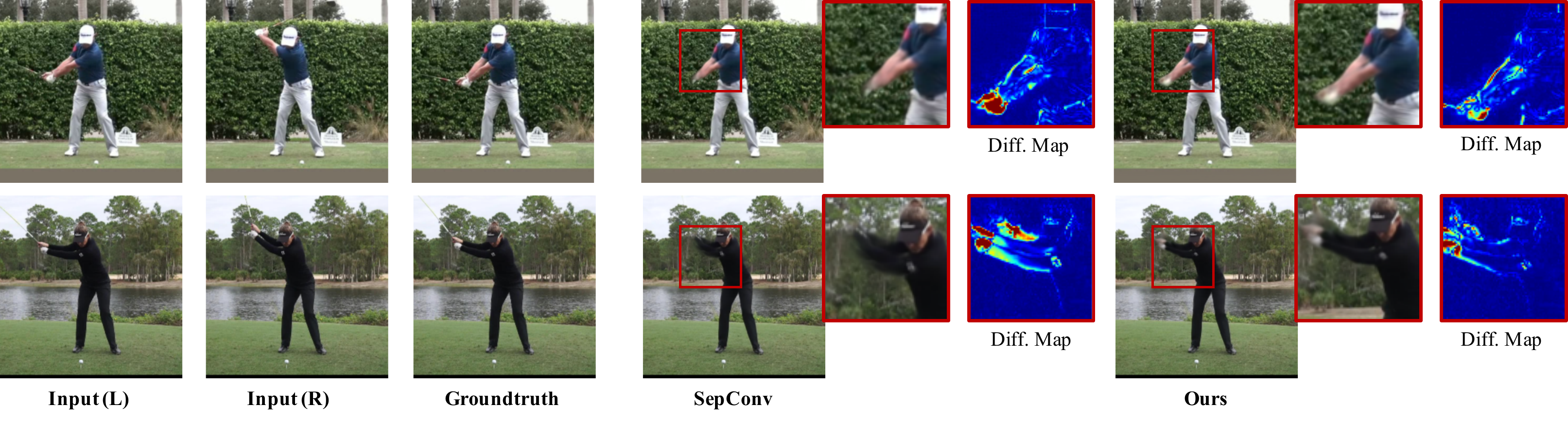}
    \caption{Qualitative results for 8x video interpolation. We observe that using program information results in uniformly higher image quality.}
    \label{fig:interpolate}
    \vspace{-5pt}
\end{figure*}
\begin{table*}
\caption{Comparison of metrics for video interpolation of 2x / 4x / 8x subsampling on the GolfDB and PoseWarp test sets. Metrics include keypoint difference in left/right elbow and wrist (lower is better), PSNR, SSIM (higher is better), and LPIPS (lower is better). 
}
    \small
    \centering\vspace{5pt}
    \begin{tabular}{llcccccccc}
        \toprule
        \multirow{2}{*}{Method} & \multirow{2}{*}{Dataset} & \multicolumn{5}{c}{Keypoint Difference$\downarrow$} &  \multicolumn{3}{c}{Perceptual Metrics} \\
        \cmidrule(lr){3-7}\cmidrule(lr){8-10}
        & & l-elbow & l-wrist & r-elbow & r-wrist & average & PSNR$\uparrow$ & SSIM$\uparrow$ & LPIPS$\downarrow$  \\
        \midrule

SepConv \cite{sepconv}  & \multirow{3}{*}{PoseWarp 2x} & 2.556  & 3.606    & 2.330    & 3.124    & 1.791 &32.285   & 0.967            & 0.023      \\
SepConv w/ heatmaps     &                & 2.357                & 3.490                 & 2.267                & 2.910                           & 1.788 &32.246               & \textbf{0.968}       & 0.025         \\
SepConv w/ programs             &        & \textbf{2.238}       & \textbf{2.982}       & \textbf{1.965}       & \textbf{2.552}              & \textbf{1.650} &\textbf{32.699}      & 0.964                & \textbf{0.021}     \\
\midrule
SepConv \cite{sepconv} & \multirow{3}{*}{GolfDB 2x} & 3.263         & 3.348          & \textbf{3.077} & 3.048     & \textbf{2.308}       & 34.860           & 0.981          & 0.017         \\
SepConv w/ heatmaps         &            & 3.395         & 3.289          & 3.211          & 2.935      & 2.355              & 35.046          & 0.982          & 0.017          \\
SepConv w/ programs          &           & \textbf{3.210} & \textbf{3.263} & 3.325          & \textbf{2.900} & 2.432 & \textbf{35.597} & \textbf{0.982} & \textbf{0.015} \\
        \midrule
SepConv \cite{sepconv} & \multirow{3}{*}{PoseWarp 4x} & 3.319                & 4.625                & 2.926                & 3.933                                    & 2.151 &\textbf{30.834}      & \textbf{0.955}       & 0.034        \\
SepConv w/ heatmaps         &            & 3.011                & 4.305                & 2.708                & 3.699                                   & 2.088 &30.82                & \textbf{0.955}       & 0.036                 \\
SepConv w/ programs         &            & \textbf{2.788}       & \textbf{3.441}       & \textbf{2.495}       & \textbf{3.068}                          & \textbf{1.938} & 30.344               & 0.951                & \textbf{0.032}   \\
        \midrule
SepConv \cite{sepconv} & \multirow{3}{*}{GolfDB 4x}  & 4.987     & 5.791      &        4.620        & 4.797                   & 3.598 & 30.488         & 0.958          & 0.035           \\
SepConv w/ heatmaps    &                 & 4.927          & 5.351          & 4.670           & 4.783               & 3.532    & 30.606         & 0.959          & 0.034          \\
SepConv w/ programs     &                & \textbf{4.466} & \textbf{4.704} & \textbf{4.085} & \textbf{4.093} & \textbf{3.091} & \textbf{30.990} & \textbf{0.959} & \textbf{0.031}  \\
\midrule
SepConv \cite{sepconv} & \multirow{3}{*}{PoseWarp 8x} & 5.520                 & 9.349                & 4.132                & 7.527                                   &  3.275 &\textbf{27.659}      & \textbf{0.923}       & 0.059           \\
SepConv w/ heatmaps         &            & 5.367                & 9.214                & 4.672                & 8.425                                   & 3.411 & 27.666               & 0.923                & 0.06               \\
SepConv w/ programs         &            & \textbf{4.371}       & \textbf{6.611}       & \textbf{3.955}       & \textbf{6.475}                          & \textbf{2.840} & 27.587               & 0.92                 & \textbf{0.054} \\
        \midrule
SepConv \cite{sepconv} & \multirow{3}{*}{GolfDB 8x}  & 10.762         & 15.852          & 8.340           & 14.075                 & 6.900 &25.672          & 0.908          & 0.071   \\
SepConv w/ heatmaps       &              & 10.914         & 16.135          & 8.839          & 15.484                   &7.317 & 25.918          & 0.913          & 0.068          \\
SepConv w/ programs        &             & \textbf{8.496} & \textbf{11.134} & \textbf{7.134} & \textbf{9.929} & \textbf{5.669} & \textbf{26.563} & \textbf{0.914} & \textbf{0.062}  \\
        \bottomrule
    \end{tabular}
    \label{tab:interpolation-number}
\vspace{-5pt}
\end{table*} 
\myparagraph{Metrics.} We use the standard PSNR, SSIM~\cite{wang2004image}, and LPIPS~\cite{zhang2018unreasonable} image metrics to estimate the fidelity of synthesized videos. We also compute the keypoint difference (KD) metric as before, but now we find the average distance of keypoints detected in synthesized video with the keypoints detected in the ground-truth video. We present KD results for the major motion joints: left elbow, left wrist, right elbow, and right wrist. %

For evaluating the quality of synthesized videos, we also perform human preference studies on Amazon Mechanical Turk (AMT). We randomly showed videos synthesized from both the baseline and our method and ask humans `Which video looks more realistic?' For each task, 30 subjects were each asked to vote on 30 random videos from the test set. 

\myparagraph{Results.} The quantitative results are shown in Table~\ref{tab:interpolation-number}. We observe that the performance of our model (SepConv w/ programs) is comparable to SepConv at 2x interpolation and improves significantly for 8x interpolation on both datasets. Our model retains higher-level structure better as indicated by the KD metrics. It also generates realistic videos as validated by the human evaluation shown in \fig{fig:interpolate-human-eval}. Figure~\ref{fig:interpolate} highlights the qualitative improvements, our model can generate more realistic hands for sparse 8x interpolation.

\subsection{Extracting Repetitive Segments}
\label{sec:exp-loop}
We now evaluate motion programs' ability on automatically extracting repetitive segments from long action clips.

\myparagraph{Data.} We use three long (15+ minutes) exercise videos from YouTube, two with cardio and one with Taichi routines. We manually annotated the videos with ground-truth labels for repetitive action segments.

\myparagraph{Results.} We generate abstract motion programs on full videos and evaluated the precision and recall of the detected for-loops. We manually annotated the videos with ground-truth labels to compute these metrics. We mark a loop as successfully extracted if the IoM (intersection over minimum) of the detected interval and ground-truth interval is greater than 0.5. The results are in Table~\ref{tab:loop-statistics}. We present samples of repetitive segments extracted in the project webpage. We also present additional qualitative examples for repetitive structure extracted from the AIST++ dance database~\cite{li2021learn,aist-dance-db}.

\myparagraph{Error analysis.} We observed a few failure cases that contribute to the less than perfect precision and recall. First, our synthesis algorithm is not good at capturing fine small motions. The problem originates with pose detectors, which have low signal-to-noise ratio for small motions.  For example, cardio00 has several small jogging motions which we do not mark as loops leading to lower recall values. Synthesis also sometimes produces poor results due to bad keypoint detection from body occlusions.  Finally, one would expect that complex 3D human actions in the $z$-plane are not well-described by 2D keypoints. We present a more detailed analysis with examples in the project webpage.

\begin{table}[t]
\caption{Comparing the number of detected segments (\# det) with ground-truth segments (\# gt) on three long exercise videos. }
\small
\setlength{\tabcolsep}{5pt}
    \centering\vspace{5pt}
    \begin{tabular}{cccccc}
        \toprule
        Video ID & Duration & \# det & \# gt & Precision & Recall \\
        \midrule
        cardio00 & 37:10 & 36 & 62 & 88.57 & 58.06 \\
        cardio01 & 17:51 & 19 & 21 & 69.35 & 90.47 \\
        taichi02 & 23:22 & 23 & 34 & 100 & 67.64 \\
        \bottomrule
    \end{tabular}
    \label{tab:loop-statistics}
    \vspace{-12pt}
\end{table}
 
\subsection{Video Prediction}
\label{sec:exp-manip}
    Once a set of for-loops has been detected in a video, motion programs enable indefinite extrapolation by executing the synthesized for-loops for additional iterations. This enables an interactive video editing platform, where one could expand or contract the repetitive section. In this section, we evaluate how well the for-loop based video prediction compare to standard baselines.
    
\myparagraph{Data.} For each long video used in the previous section, we extract a dataset of all the detected for-loops. We use an 80-20 train-test split. 

\myparagraph{Setup.} For each dataset, we use the training set to train our baselines, HierchVid~\cite{villegas17hierchvid} and HistryRep~\cite{wei2020his}. HierchVid is an LSTM-based future pose predictor while HistryRep uses an attention-based model. At test time, we use the first half of the videos as input and compare the future poses generated with the second half as ground-truth. We also use first half as input to generate a motion program with for-loop and unroll it indefinitely to generate future poses. We then synthesize future frames from poses for both the methods using the Everybody Dance Now~\cite{chan2019dance} GAN and compare it with the ground-truth frames.

\begin{figure}[t]
    \centering
    \includegraphics[width=.49\linewidth]{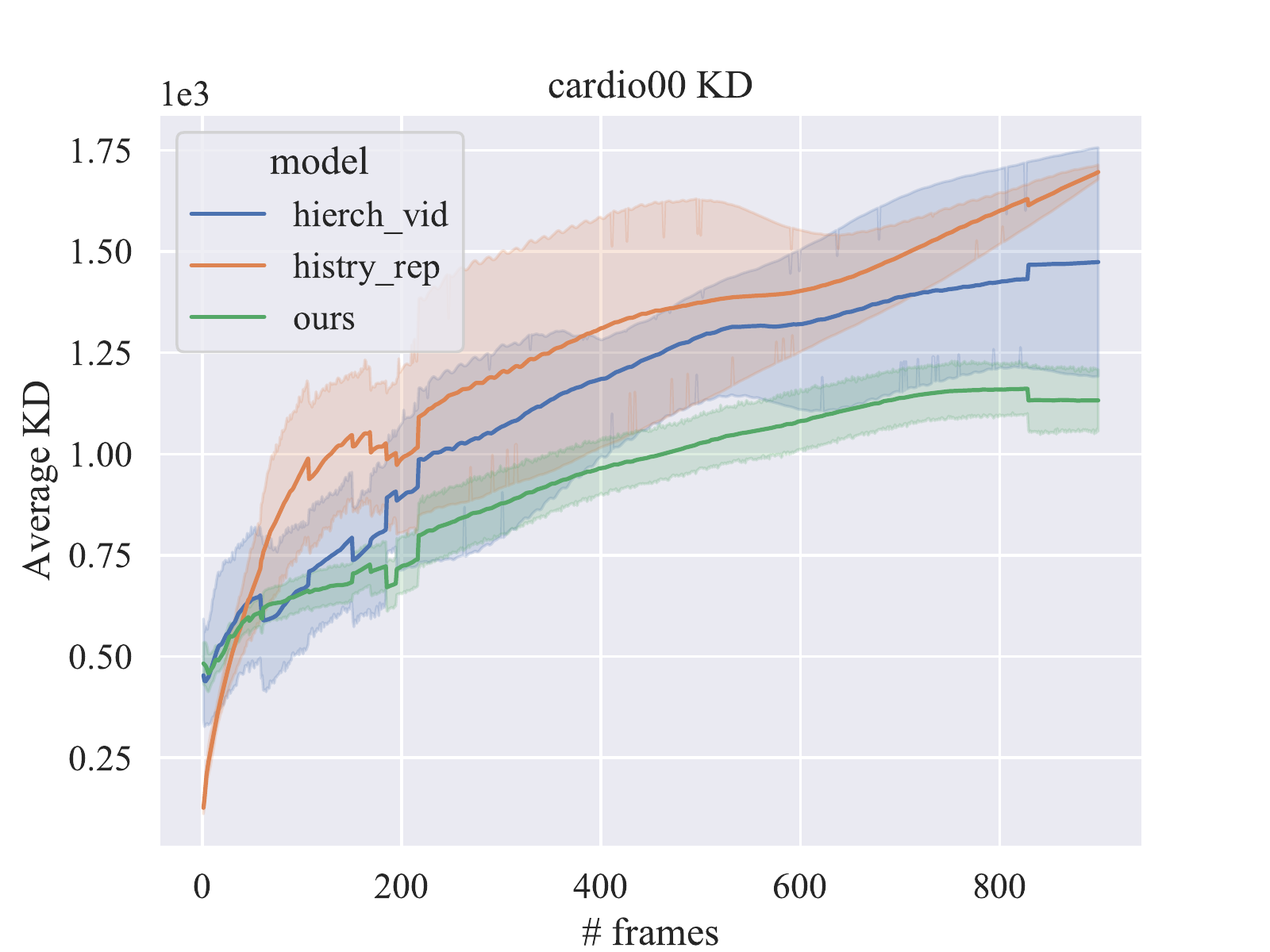}
    \includegraphics[width=.49\linewidth]{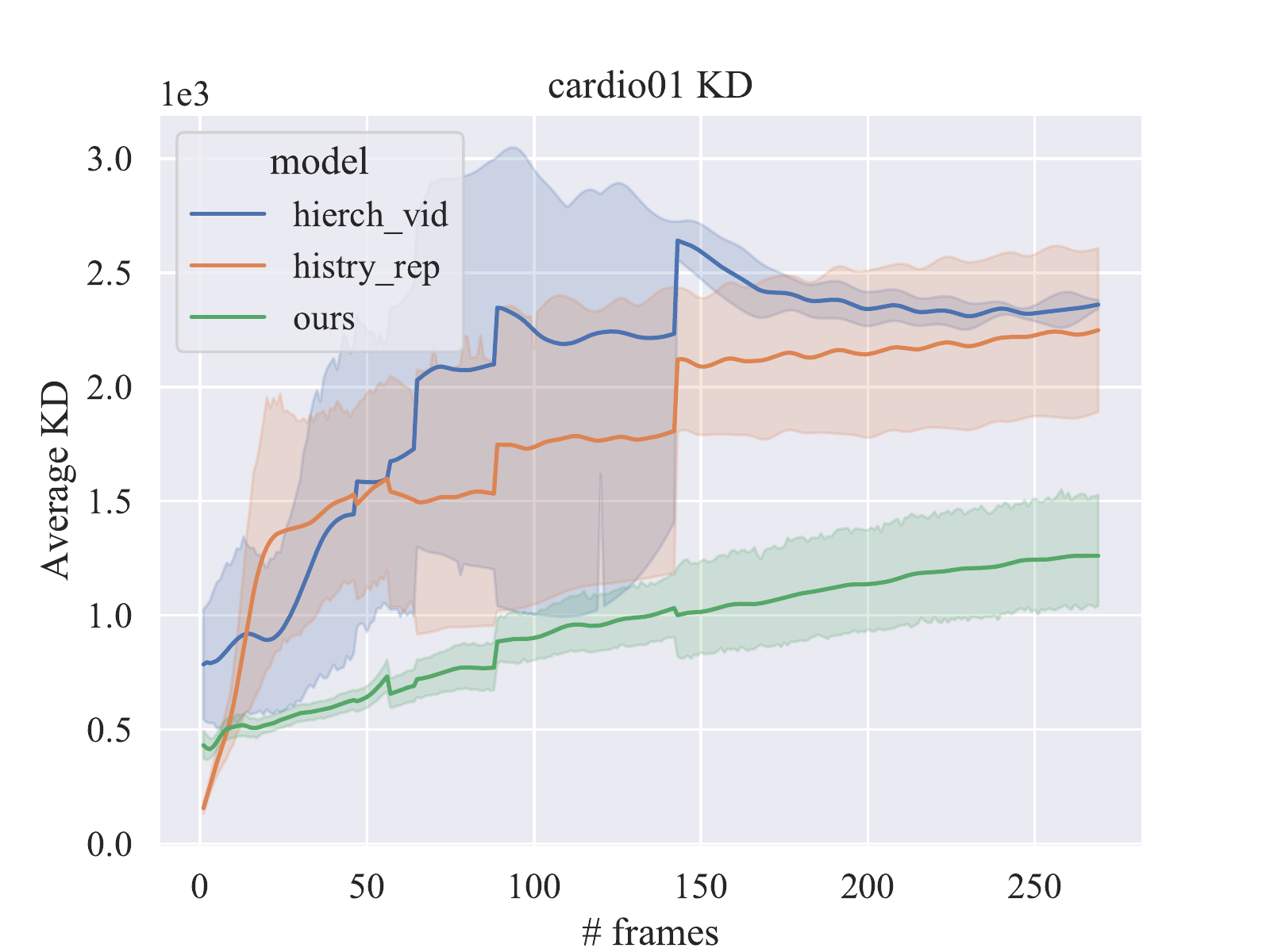}
    \\%
    \vspace{-5pt}
    \includegraphics[width=.49\linewidth]{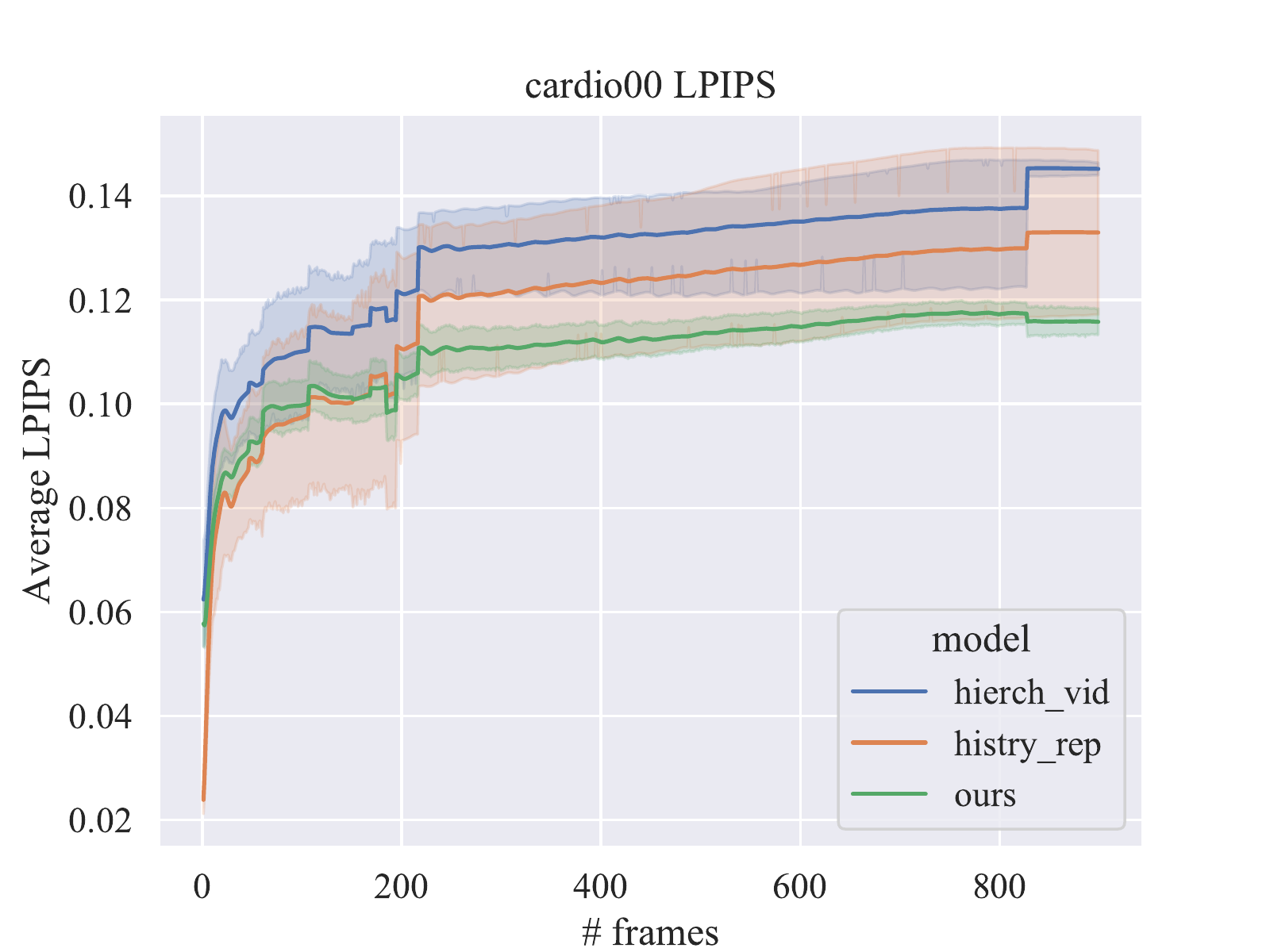}
    \includegraphics[width=.49\linewidth]{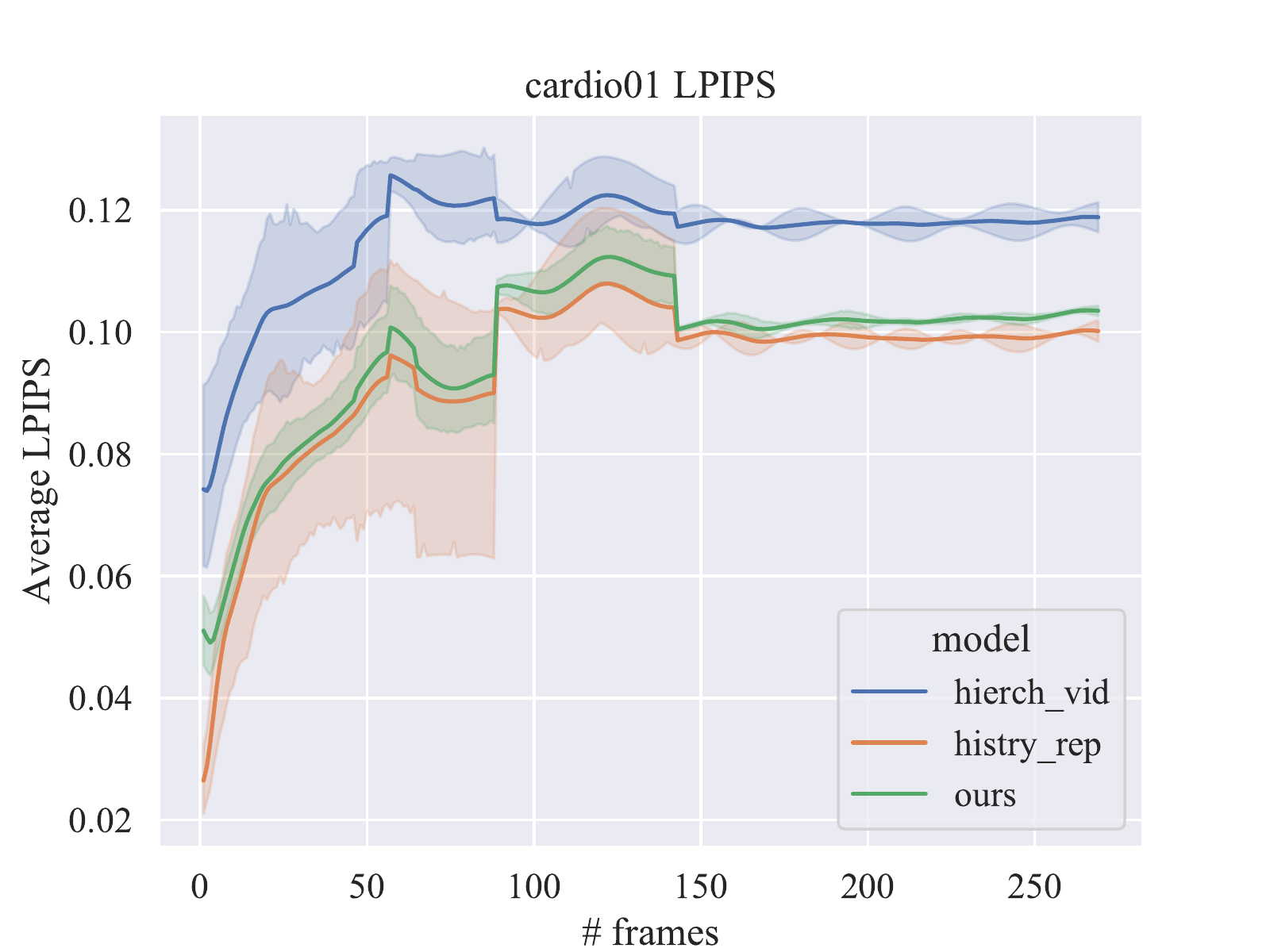}
    \vspace{-2pt}
    \caption{Comparison of HierchVid~\cite{villegas17hierchvid} (blue), HistryRep~\cite{wei2020his} (orange)  and ours (green) on video prediction. Top row: KD$\downarrow$ (keypoint difference), bottom row: LPIPS$\downarrow$ on loops extracted from cardio00 and cardio01.}
    \vspace{-10pt}
    \label{fig:extrapolate-plots}
\end{figure}

\begin{figure}[t]
    \centering
    \includegraphics[width=1\linewidth]{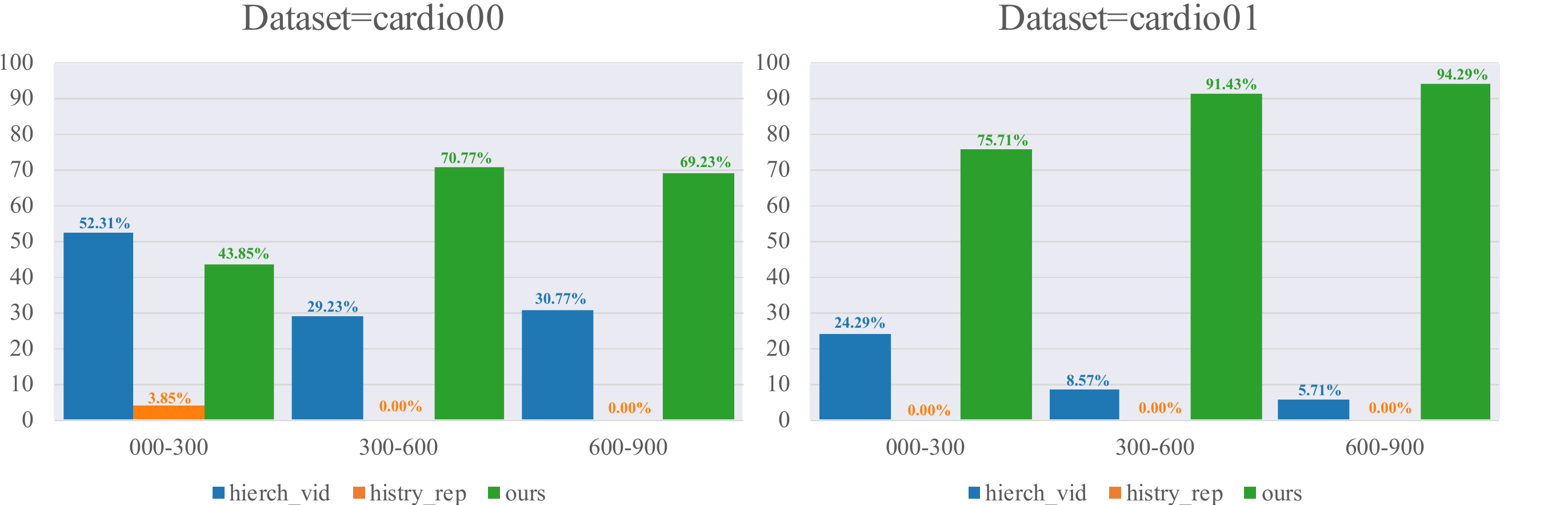}
    \caption{Human evaluation of video prediction across different future frame slices on two datasets. We observe that extrapolating via motion programs leads to more consistent future prediction of repetitive segments}
    \label{fig:extrarpolate-human-eval}
    \vspace{-10pt}
\end{figure}

\myparagraph{Metrics.} Similar to video interpolation, we use the PSNR, SSIM and LPIPS image metrics when ground truth frames are available. We also compute the keypoint difference (KD) with the ground-truth poses. We average our results over 10 runs. Since predicting future frames is a stochastic task and direct comparison to ground-truth might be an incomplete analysis, we also perform human preference studies on AMT. We randomly showed videos synthesized from both the methods and ask humans `Which video looks more similar to the ground-truth?'. For each dataset, 10 subjects were each asked to vote for all the videos from the test set. To have a granular understanding, we did three studies, comparing the first 900 frames in sets of 300 frames.

\myparagraph{Results.} We present quantitative results in Figure~\ref{fig:extrapolate-plots} where we plot the average KD and LPIPS metrics on the two datasets against the number of future frames predicted. We observe that as the number of future frames increases, the gap between the baselines and our model increases. This is due to the fact that both HierchVid and HistryRep perform sequence modelling and do not have higher level knowledge of repetitions. After the first few frames, the error compounds and the model diverges to performing a different action. This fact is reinforced by the human evaluation study, results of which are in Figure~\ref{fig:extrarpolate-human-eval}. On the cardio00 dataset, HierchVid does well on the first few frames, though in the later frames our model is more consistent with the ground-truth. We present additional results on more metrics on our webpage.

\section{Discussion}
We have presented a hierarchical motion understanding framework based on motion programs. From the input video, we can automatically build primitive-level understanding while also capturing higher-level repetitive structures. Experiments show that these programs can capture a wide range of semantically meaningful actions, enhance  human motion understanding, and enable applications such as video synthesis for computer animation. Like all other visual content generation methods, we urge users of our models to be aware of potential ethical and societal concerns and to apply them with good intent. %

\myparagraph{Acknowledgements.} 
This work is in part supported by Magic Grant from the Brown Institute for Media Innovation, the Samsung Global Research Outreach (GRO) Program, Autodesk, Amazon Web Services, and Stanford HAI for AWS Cloud Credits.

{\small
\bibliography{motionprog,bpi}

\begin{thebibliography}{10}\itemsep=-1pt

\bibitem{aristidou2018deep}
Andreas Aristidou, Daniel Cohen-Or, Jessica~K Hodgins, Yiorgos Chrysanthou, and
  Ariel Shamir.
\newblock Deep motifs and motion signatures.
\newblock {\em ACM TOG}, 2018.

\bibitem{posewarp}
Guha Balakrishnan, Amy Zhao, Adrian~V Dalca, Fredo Durand, and John Guttag.
\newblock Synthesizing images of humans in unseen poses.
\newblock In {\em CVPR}, 2018.

\bibitem{bullock2006least}
Randy Bullock.
\newblock Least-squares circle fit.
\newblock {\em Developmental Testbed Center}, 2006.

\bibitem{openpose}
Zhe Cao, Gines Hidalgo, Tomas Simon, Shih-En Wei, and Yaser Sheikh.
\newblock Openpose: Realtime multi-person 2d pose estimation using part
  affinity fields.
\newblock {\em IEEE TPAMI}, 2019.

\bibitem{chan2019dance}
Caroline Chan, Shiry Ginosar, Tinghui Zhou, and Alexei~A Efros.
\newblock Everybody dance now.
\newblock In {\em ICCV}, 2019.

\bibitem{davis2018visual}
Abe Davis and Maneesh Agrawala.
\newblock Visual rhythm and beat.
\newblock {\em ACM TOG}, 2018.

\bibitem{denton2018stochastic}
Emily Denton and Rob Fergus.
\newblock {Stochastic Video Generation with a Learned Prior}.
\newblock In {\em ICML}, 2018.

\bibitem{ellis2018learning}
Kevin Ellis, Daniel Ritchie, Armando Solar-Lezama, and Josh Tenenbaum.
\newblock {Learning to Infer Graphics Programs from Hand-{Drawn} Images}.
\newblock In {\em NeurIPS}, 2018.

\bibitem{alphapose}
Hao-Shu Fang, Shuqin Xie, Yu-Wing Tai, and Cewu Lu.
\newblock {RMPE}: Regional multi-person pose estimation.
\newblock In {\em ICCV}, 2017.

\bibitem{gleicher1997motion}
Michael Gleicher.
\newblock Motion editing with spacetime constraints.
\newblock In {\em I3D}, 1997.

\bibitem{gleicher1998constraint}
Michael Gleicher and Peter Litwinowicz.
\newblock Constraint-based motion adaptation.
\newblock {\em The journal of visualization and computer animation}, 1998.

\bibitem{jiang2018super}
Huaizu Jiang, Deqing Sun, Varun Jampani, Ming-Hsuan Yang, Erik Learned-Miller,
  and Jan Kautz.
\newblock Super slomo: High quality estimation of multiple intermediate frames
  for video interpolation.
\newblock In {\em CVPR}, 2018.

\bibitem{kovar2008motion}
Lucas Kovar, Michael Gleicher, and Fr{\'e}d{\'e}ric Pighin.
\newblock Motion graphs.
\newblock In {\em SIGGRAPH}, 2008.

\bibitem{li2017grass}
Jun Li, Kai Xu, Siddhartha Chaudhuri, Ersin Yumer, Hao Zhang, and Leonidas
  Guibas.
\newblock {{GRASS}: Generative Recursive Autoencoders for Shape Structures}.
\newblock {\em ACM TOG}, 2017.

\bibitem{li2019grains}
Manyi Li, Akshay~Gadi Patil, Kai Xu, Siddhartha Chaudhuri, Owais Khan, Ariel
  Shamir, Changhe Tu, Baoquan Chen, Daniel Cohen-Or, and Hao Zhang.
\newblock {{GRAINS}: Generative Recursive Autoencoders for {INdoor} Scenes}.
\newblock {\em ACM TOG}, 2019.

\bibitem{li2021learn}
Ruilong Li, Shan Yang, David~A. Ross, and Angjoo Kanazawa.
\newblock Learn to dance with aist++: Music conditioned 3d dance generation,
  2021.

\bibitem{lindenmayer1968mathematical}
Aristid Lindenmayer.
\newblock Mathematical models for cellular interactions in development ii.
  simple and branching filaments with two-sided inputs.
\newblock {\em Journal of theoretical biology}, 1968.

\bibitem{liu2019deep}
Yu-Lun Liu, Yi-Tung Liao, Yen-Yu Lin, and Yung-Yu Chuang.
\newblock Deep video frame interpolation using cyclic frame generation.
\newblock In {\em AAAI}, volume~33, 2019.

\bibitem{liu2017video}
Ziwei Liu, Raymond Yeh, Xiaoou Tang, Yiming Liu, and Aseem Agarwala.
\newblock {Video Frame Synthesis Using Deep Voxel Flow}.
\newblock In {\em ICCV}, 2017.

\bibitem{lu2020layered}
Erika Lu, Forrester Cole, Tali Dekel, Weidi Xie, Andrew Zisserman, David
  Salesin, William~T Freeman, and Michael Rubinstein.
\newblock Layered neural rendering for retiming people in video.
\newblock {\em SIGGRAPH Asia}, 2020.

\bibitem{mao2019program}
Jiayuan Mao, Xiuming Zhang, Yikai Li, William~T. Freeman, Joshua~B. Tenenbaum,
  and Jiajun Wu.
\newblock {Program-{Guided} Image Manipulators}.
\newblock In {\em ICCV}, 2019.

\bibitem{golfdb}
William McNally, Kanav Vats, Tyler Pinto, Chris Dulhanty, John McPhee, and
  Alexander Wong.
\newblock Golfdb: A video database for golf swing sequencing.
\newblock In {\em CVPR Workshop}, June 2019.

\bibitem{sepconv}
Simon Niklaus, Long Mai, and Feng Liu.
\newblock Video frame interpolation via adaptive separable convolution.
\newblock In {\em CVPR}, 2017.

\bibitem{niu2018im2struct}
Chengjie Niu, Jun Li, and Kai Xu.
\newblock {{Im2Struct}: Recovering 3d Shape Structure from a Single Rgb Image}.
\newblock In {\em CVPR}, 2018.

\bibitem{rose1999verbs}
Charles~F Rose, Bobby Bodenheimer, and Michael~F Cohen.
\newblock {\em Verbs and adverbs: Multidimensional motion interpolation using
  radial basis functions}.
\newblock Citeseer, 1999.

\bibitem{sharma2018csgnet}
Gopal Sharma, Rishabh Goyal, Difan Liu, Evangelos Kalogerakis, and Subhransu
  Maji.
\newblock {{CSGNet}: Neural Shape Parser for Constructive Solid Geometry}.
\newblock In {\em CVPR}, 2018.

\bibitem{tian2019learning}
Yonglong Tian, Andrew Luo, Xingyuan Sun, Kevin Ellis, William~T. Freeman,
  Joshua~B. Tenenbaum, and Jiajun Wu.
\newblock {Learning to Infer and Execute {3D} Shape Programs}.
\newblock In {\em ICLR}, 2019.

\bibitem{aist-dance-db}
Shuhei Tsuchida, Satoru Fukayama, Masahiro Hamasaki, and Masataka Goto.
\newblock Aist dance video database: Multi-genre, multi-dancer, and
  multi-camera database for dance information processing.
\newblock In {\em ISMIR}, pages 501--510, Delft, Netherlands, Nov. 2019.

\bibitem{villegas17hierchvid}
Ruben Villegas, Jimei Yang, Yuliang Zou, Sungryull Sohn, Xunyu Lin, and Honglak
  Lee.
\newblock {Learning to Generate Long-term Future via Hierarchical Prediction}.
\newblock In {\em ICML}, 2017.

\bibitem{wang2011symmetry}
Yanzhen Wang, Kai Xu, Jun Li, Hao Zhang, Ariel Shamir, Ligang Liu, Zhiquan
  Cheng, and Yueshan Xiong.
\newblock {Symmetry Hierarchy of Man-{Made} Objects}.
\newblock {\em CGF}, 2011.

\bibitem{wang2004image}
Zhou Wang, Alan~C. Bovik, Hamid~R. Sheikh, and Eero~P. Simoncelli.
\newblock {Image Quality Assessment: From Error Visibility to Structural
  Similarity}.
\newblock {\em IEEE TIP}, 2004.

\bibitem{wei2020his}
Mao Wei, Liu Miaomiao, and Salzemann Mathieu.
\newblock History repeats itself: Human motion prediction via motion attention.
\newblock In {\em ECCV}, 2020.

\bibitem{young2019learning}
Halley Young, Osbert Bastani, and Mayur Naik.
\newblock {Learning Neurosymbolic Generative Models via Program Synthesis}.
\newblock In {\em ICML}, 2019.

\bibitem{zhang2018unreasonable}
Richard Zhang, Phillip Isola, Alexei~A. Efros, Eli Shechtman, and Oliver Wang.
\newblock {The Unreasonable Effectiveness of Deep Networks As a Perceptual
  Metric}.
\newblock In {\em CVPR}, 2018.

\end{thebibliography}
\bibliographystyle{ieee_fullname}
}

\end{document}